\documentclass[conference]{IEEEtran}
\IEEEoverridecommandlockouts
\usepackage{cite}
\usepackage{amsmath,amssymb,amsfonts}
\usepackage{algorithmic}
\usepackage{graphicx}
\usepackage{textcomp}
\usepackage{xcolor}
\usepackage{booktabs}
\usepackage{multirow}
\usepackage{url}
\usepackage[hidelinks]{hyperref}
\usepackage[english]{babel}
\usepackage[T1]{fontenc}

\def\BibTeX{{\rm B\kern-.05em{\sc i\kern-.025em b}\kern-.08em
    T\kern-.1667em\lower.7ex\hbox{E}\kern-.125emX}}

\begin{document}

\title{Autoencoder Architectures for Athlete Performance Scoring from Wearable Telemetry}

\author{\IEEEauthorblockN{Mateusz Kubita}
\IEEEauthorblockA{\textit{Student of }\\
\textit{Warsaw University of Technology} \\
Warsaw, Poland \\
mateusz.kubita.stud@pw.edu.pl}
\and
\IEEEauthorblockN{Jan Zubalewicz}
\IEEEauthorblockA{\textit{Student of}\\
\textit{Warsaw University of Technology} \\
Warsaw, Poland \\
jan.zubalewicz.stud@pw.edu.pl}
\and
\IEEEauthorblockN{Krzysztof Siwek}
\IEEEauthorblockA{\textit{Warsaw University of Technology} \\
Warsaw, Poland \\
krzysztof.siwek@pw.edu.pl}
}

\maketitle

\begin{abstract}
Wearable devices produce large, high dimensional training logs for
everyday runners, and interpretation rather than data collection is now
the limiting step. This paper evaluates five dimensionality reduction
models, three autoencoder variants, PCA, and a Variational Autoencoder,
on their ability to compress nine sensor runner profiles into a single
scalar performance indicator, the latent score. Because the
setting is fully unsupervised, model quality is assessed along two
complementary axes: reconstruction error (Mean Squared Error) and
latent score interpretability, measured via Spearman and Kendall rank
correlations, Mutual Information, and Permutation Importance. These are
combined into a composite selection criterion that prevents selecting
models on reconstruction accuracy alone. Feature rankings from the four
metrics are aggregated via a modified Borda count, and their stability
is confirmed by bootstrap validation. A two feature linear baseline is
included to anchor the comparison. Deep autoencoder achieved the lowest
reconstruction error and the highest composite score. Once the
PCA hidden layers were widened, the deeper variants
became closely competitive with Deep AE on the composite criterion,
indicating that the limiting factor was hidden layer capacity rather
than the one dimensional bottleneck. Running pace, aerobic decoupling,
and average heart rate emerged as the dominant latent score drivers
across all models and resampling runs, consistent with established
physiology. \end{abstract}


\section{Introduction}

Monitoring physical activity through IoT systems produces dense
datasets combining kinematic, physiological, and environmental
variables \cite{b1}. A single session recorded by a smartwatch and
chest strap can generate dozens of overlapping signals, and continuous
heart rate monitoring, GPS pacing, and ground contact time measurement
are now standard even outside elite sport \cite{b10}. The practical
challenge has shifted from data collection to synthesis. Coaches need
a compact indicator that reflects performance without requiring manual
inspection of multiple plots. During marathon preparation each athlete
accumulates thousands of data points. In practice coaches review these
through hand crafted summaries and fixed weights, an approach that is
easy to apply but that can obscure nonlinear interactions between
workload, fatigue, and cardiovascular response \cite{b6}. Uphill
running, for example, lowers pace while raising heart rate
simultaneously, a pattern that reflects terrain rather than reduced
fitness. Simple weighted averages cannot capture this kind of context
reliably.

Dimensionality reduction techniques such as PCA are widely used in
sports analytics, but their ability to represent complex physiological
dynamics is constrained by linearity \cite{b18}. Existing unsupervised
approaches typically optimize reconstruction accuracy alone, without
checking whether the derived score is interpretable or stable across
data splits. No widely accepted framework simultaneously compresses
multi sensor telemetry into a single interpretable score, preserves
nonlinear physiological relationships, and provides stable rankings. This
work addresses that gap with a unified evaluation framework for
unsupervised athlete ranking. The specific contributions are: reduction
of each athlete profile to a one dimensional latent score, model
selection based on a composite criterion combining reconstruction
accuracy and ranking interpretability, feature level interpretation
using four complementary Explainable AI metrics aggregated by a
modified Borda count; and bootstrap based ranking stability assessment
with confidence intervals.

\section{Related Work}

Autoencoders have been used for nonlinear dimensionality reduction
since Hinton and Salakhutdinov showed that a deep network can compress
data more effectively than PCA when the underlying structure is
nonlinear \cite{b3}. Variational autoencoders extended this idea with a
probabilistic latent space \cite{b4}, which makes them attractive for
noisy sensor data. In sport and health, wearable telemetry has been
modeled with both classical and deep methods \cite{b6,b10}. Most of
the applications either keep the latent space high dimensional or use
it only for clustering, not for a single interpretable ranking score.
Machine learning is now widely applied to sport specific tasks, from
movement recognition to injury forecasting \cite{b1,b17}. Reviews of
the area note that model performance is often reported without much
attention to interpretability \cite{b17}. Explainability methods such
as permutation importance and additive feature attribution
\cite{b5,b9,b12} have been used to open up supervised models, yet they
are rarely combined into a consensus procedure for an unsupervised
score. Rank aggregation offers a principled way to merge several
importance measures \cite{b2}, and bootstrap resampling provides
stability estimates \cite{b14}. The framework in this paper brings
these threads together. It compresses telemetry to one dimension, ranks
the drivers of that dimension with four metrics, and tests their
stability across resamples.

\section{Data and Processing Pipeline}

\subsection{Dataset Characteristics}

The dataset originates from exports of the Golden Cheetah platform
and is included in the project repository as \texttt{data/activities.csv}.
It was obtained from a publicly available export and no restricted
identifiers were used; all records were analyzed in anonymized form.
Because the data were publicly accessible and anonymized at the time of
download, no institutional review board approval was required.
Real sessions include dropped GPS samples, temporary heart rate
disconnects, and recording artifacts such as watches left running after
training, so a dedicated preprocessing stage was required. The cleaning
process covered removing records with missing values, handling extreme
tracking errors, and applying outlier filtration. After preprocessing
the final dataset comprised \textbf{45,836 observations}. A formal
80/10/10 train, validation, and test split was created during
preprocessing and applied consistently throughout the experiments.
Each profile was described by nine features extracted from the raw
activity files: pace (\textit{pace\_min\_km}), average heart rate
(\textit{average\_hr}), elevation gain (\textit{elevation\_gain}),
total distance (\textit{total\_distance}), cadence
(\textit{final\_cadence}), aerobic decoupling
(\textit{aerobic\_decoupling}), age (\textit{age}), body weight
(\textit{athlete\_weight}), and a binary sex indicator
(\textit{is\_male}). The binary treatment of sex does not account for
hormonal or anatomical variation within the population and is an
acknowledged simplification. Aerobic decoupling quantifies the drift in
the pace to heart rate ratio between the first and second halves of a
session \cite{b11}. High values indicate fatigue, dehydration, or
insufficient aerobic conditioning, making it an informative marker for
session quality and longitudinal fitness trends. All variables were scaled
with Min Max normalization on the training split.

\subsection{Compared Model Architectures}

Five architectures were evaluated, each reducing the nine dimensional
input to a one dimensional bottleneck. The \textbf{Simple AE} applies
a direct single layer mapping and serves as a baseline for neural
compression. The \textbf{Medium AE} adds one hidden layer on each side
of the bottleneck, implementing the topology
Input $\rightarrow$ 16 $\rightarrow$ 1 $\rightarrow$ 16 $\rightarrow$
Output \cite{b7}. The \textbf{Deep AE} is a multilayer structure with
topology Input $\rightarrow$ 32 $\rightarrow$ 8 $\rightarrow$ 1
$\rightarrow$ 8 $\rightarrow$ 32 $\rightarrow$ Output \cite{b3}. A
deliberate design choice separates the compression target from the
representational capacity. The bottleneck is fixed at one dimension
because the downstream task is a single scalar ranking score, so the
latent representation must be one dimensional to be interpretable as a
performance ordering. The surrounding hidden layers, by contrast, are
kept generous (16, 32, then 8) so that the encoder has enough
capacity to learn a useful nonlinear mapping before it reaches the
bottleneck. Narrow hidden layers would compress the input too early and
starve the model, which is what limited earlier deterministic variants.
With about 36,000 training samples for nine features and early stopping
in place, these wider layers do not raise a meaningful overfitting risk.
For context, the first principal component explains approximately 46.4\%
of the variance, indicating that one linear dimension captures a
substantial but incomplete fraction of the dataset structure. A
systematic comparison of $k = 1, 2, 3$ bottleneck dimensionalities is
left for future work. \textbf{PCA} provides a strictly linear baseline
\cite{b18}. The \textbf{VAE} uses a probabilistic encoder with an
intermediate dimension of 32, regularizing the latent space by jointly
optimizing reconstruction loss and Kullback-Leibler divergence \cite{b4},
which encourages similar athlete profiles to cluster in the latent space.

\section{Experimental Protocol}

Hyperparameters for all neural architectures were tuned on the
validation set. Early stopping was applied based on validation loss
with a patience of ten epochs. Each neural model was trained five times
with different random seeds, and the reported metrics are averages
across those runs to reduce sensitivity to initialization variance.
PCA does not require stochastic training and was fitted once on the
training split. Model selection used a composite criterion that
incorporates both reconstruction accuracy and latent score
interpretability,
\begin{equation}
    \text{Score}_{\text{sel}} = \alpha \cdot (1 - \text{MSE}_{\text{norm}})
    + (1 - \alpha) \cdot Q
    \label{eq:composite}
\end{equation}
where $\text{MSE}_{\text{norm}}$ is the min-max normalized
reconstruction error across candidate models, $Q$ is the Latent Score
Quality defined in Section~\ref{sec:interpretation}, and $\alpha = 0.5$
assigns equal weight to both criteria. This formulation avoids
selecting models purely on reconstruction accuracy, which would favor
linear projections. All evaluation metrics in Section~\ref{sec:results}
was computed on the held out test set after hyperparameter tuning.
Table~\ref{tab:composite_derivation} shows the intermediate values
used to compute the composite selection score in
Equation~(\ref{eq:composite}). The observed minimum and maximum
test set MSE values are $\min(\text{MSE}) = 0.001781$ (Deep AE) and
$\max(\text{MSE}) = 0.169234$ (Simple AE).

\begin{table*}[htb]
    \caption{Derivation of the composite selection score. For each
    model we show the test MSE, the min-max normalized MSE,
    the complement $1-\text{MSE}_{\text{norm}}$, the Latent Score
    Quality $Q$, and the final $\text{Score}_{\text{sel}}$ with
    $\alpha = 0.5$.}
    \begin{center}
    \begin{tabular}{lccccc}
        \toprule
        \textbf{Model} & \textbf{MSE} & \textbf{$\text{MSE}_{\text{norm}}$} &
        \textbf{$1-\text{MSE}_{\text{norm}}$} & \textbf{$Q$} &
        \textbf{$\text{Score}_{\text{sel}}$} \\
        \midrule
        Simple AE        & 0.169234 & 1.000000 & 0.000000 & 0.814000 & 0.407000 \\
        Medium AE        & 0.006123 & 0.025930 & 0.974070 & 0.810000 & 0.892035 \\
        \textbf{Deep AE} & \textbf{0.001781} & \textbf{0.008563} & \textbf{0.999882} & \textbf{0.943000} & \textbf{0.971500} \\
        PCA              & 0.002354 & 0.003422 & 0.996578 & 0.858000 & 0.927289 \\
        VAE              & 0.004352 & 0.015354 & 0.984646 & 0.845000 & 0.914823 \\
        \bottomrule
    \end{tabular}
    \label{tab:composite_derivation}
    \end{center}
\end{table*}

The composite criterion uses an equal weighting ($\alpha = 0.5$) by
convention, so we report how model selection responds to that choice.
Table~\ref{tab:alpha_sensitivity} lists the composite score of every
model at $\alpha = 0.3$, $0.5$, and $0.7$. Deep AE ranks first across all
three settings, and the ordering of the leading models is stable. This
indicates that the conclusion does not hinge on the exact weight.

\begin{table}[htbp]
    \caption{Composite score by weighting $\alpha$.}
    \begin{center}
    \begin{tabular}{lccc}
        \toprule
        \textbf{Model} & \textbf{$\alpha=0.3$} & \textbf{$\alpha=0.5$} &
        \textbf{$\alpha=0.7$} \\
        \midrule
        Simple AE        & 0.570 & 0.407 & 0.244 \\
        Medium AE        & 0.859 & 0.892 & 0.925 \\
        \textbf{Deep AE} & \textbf{0.960} & \textbf{0.972} & \textbf{0.983} \\
        PCA              & 0.900 & \textbf{0.927} & 0.955 \\
        VAE              & 0.887 & 0.915 & 0.943 \\
        \bottomrule
    \end{tabular}
    \label{tab:alpha_sensitivity}
    \end{center}
\end{table}

\section{Feature Interpretation Methodology}
\label{sec:interpretation}

Explaining unsupervised latent representations is a recognized
research challenge \cite{b5}. In coaching contexts, a lower latent
score should be traceable to a specific variable, such as elevated
heart rate at a given pace or a cadence shift consistent with fatigue,
rather than emerging from an opaque combination of inputs. To reduce
over interpretation from any single statistical perspective this study
adopts a multimetric consensus procedure.

\subsection{Separate Analysis and Consensus}

Feature rankings were computed with four independent metrics, each
capturing a different facet of the variable score relationship.
Spearman rank correlation captures monotonic relationships. Kendall
rank correlation ($\tau$) is more resistant to outliers, which is
valuable for wearable derived features. Mutual Information identifies
nonlinear dependencies that correlations may miss. Permutation
Importance, computed using a Random Forest Regressor with 80
estimators, evaluates how much shuffling a feature degrades predictive
accuracy \cite{b9}, providing a model based complement to the
statistical measures. Individual rankings were integrated using a
modified Borda count \cite{b2}. The normalized score for feature $f$
under metric $m$ is
\begin{equation}
    s_{m,f} = 1 - \frac{r_{m,f}-1}{p-1}
\end{equation}
where $r_{m,f}$ is the rank of feature $f$ under metric $m$ and $p$
is the total number of features. The final Combined Impact Score is
\begin{equation}
    S_f = \sum_m w_m s_{m,f}, \quad \text{where} \quad \sum_m w_m = 1
\end{equation}
Default weights were 0.25 for Spearman, 0.20 for Kendall, 0.25 for
Mutual Information, and 0.30 for Permutation Importance. The
upweighting of Permutation Importance reflects its direct connection
to predictive degradation. Sensitivity to this weight assignment was
assessed by sampling 1,000 weight vectors from a Dirichlet distribution.
The reference top three set \{pace, average heart rate, aerobic
decoupling\} was preserved in 639 of 1,000 samples (63.9\%) when order
was ignored, and in 68 samples (6.8\%) when exact ordering was required.
This indicates moderate robustness: the dominant features do not depend
on the specific default weights, but neither are they completely
insensitive to moderate perturbations. Table~\ref{tab:dirichlet_summary}
summarizes these results.

\begin{table}[htbp]
    \caption{Dirichlet sensitivity summary (1,000 weight samples).}
    \begin{center}
    \begin{tabular}{lrr}
        \toprule
        \textbf{Measure} & \textbf{Count} & \textbf{Percent (\%)} \\
        \midrule
        Top 3 preserved (order agnostic) & 639 & 63.9 \\
        Exact top 3 ordering preserved   &  68 &  6.8 \\
        \bottomrule
    \end{tabular}
    \label{tab:dirichlet_summary}
    \end{center}
\end{table}

\begin{figure}[htbp]
    \centerline{\includegraphics[width=\columnwidth]{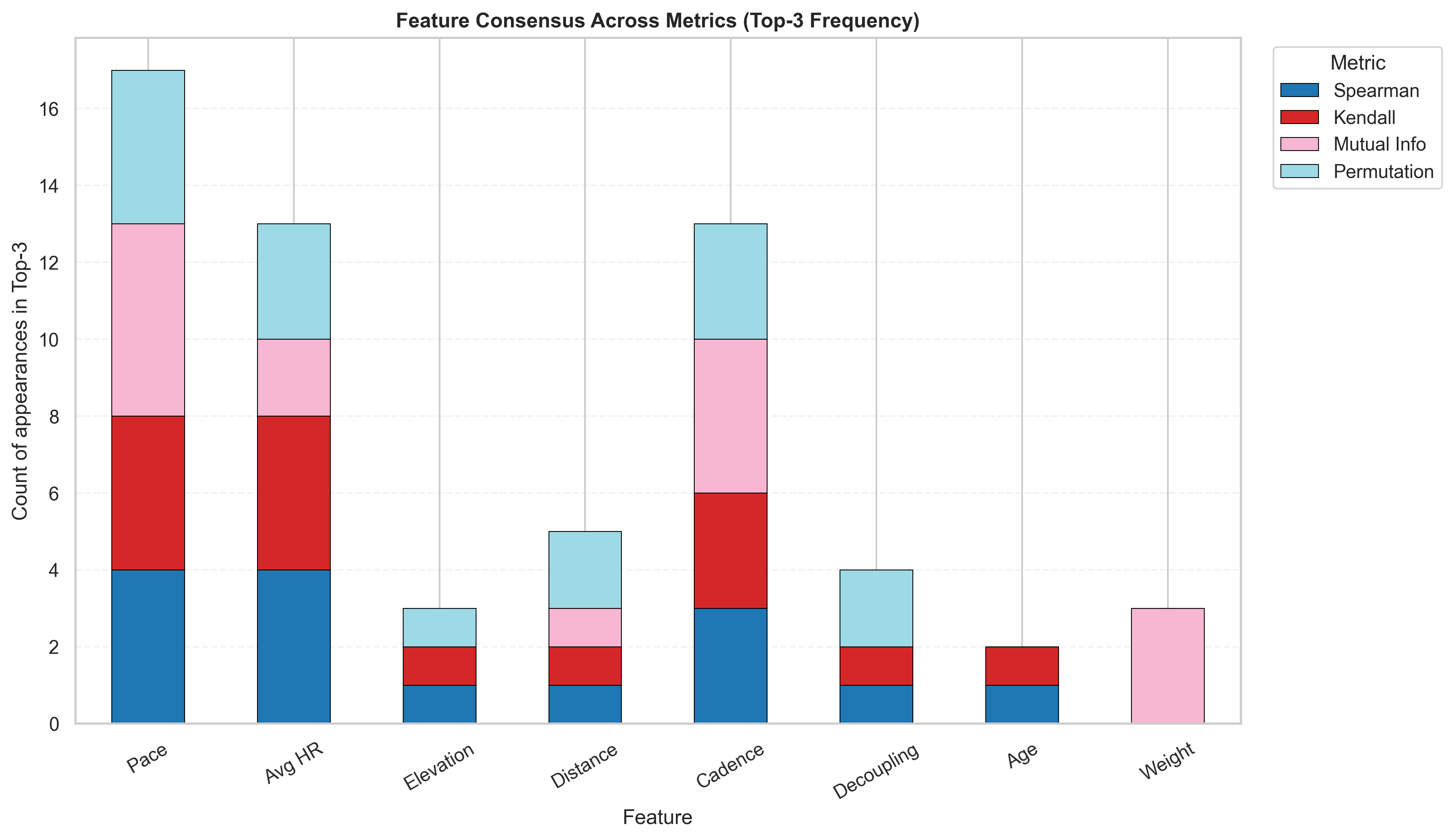}}
    \caption{Top 3 feature inclusion frequency across the four
    importance metrics. Stacked bars show how consistently each feature appears among the three most important variables under Spearman, Kendall, Mutual Information, and Permutation Importance.}
    \label{fig:metricwise_consensus}
\end{figure}

The Latent Score Quality $Q$ in Equation~(\ref{eq:composite}) is the
mean normalized Combined Impact Score of the top ranked feature,
averaged across the four metrics. Higher values indicate that the top
feature receives consistently high rankings, reflecting a more
interpretable and stable latent representation.
Figure~\ref{fig:metricwise_consensus} shows how often each feature
appears in the top three across the four metrics.

\subsection{Stability and Bootstrap Validation}

To determine whether the identified key features reflect genuine
physiological signal rather than artifacts of a particular data split,
bootstrap validation with replacement was applied \cite{b14}. For each
feature the analysis reports the top 3 inclusion probability, the
median rank, and the 95\% confidence interval of the rank distribution
across resamples. A probability of 1.000 with a zero width confidence
interval indicates that the feature appeared in the top three in every
resample, representing the strongest evidence of stability within this
framework.

\section{Results}
\label{sec:results}

\subsection{Model Comparison and Reconstruction Quality}

Table~\ref{tab:model_comparison} summarizes the test results for
all five architectures. Deep AE achieved the lowest MSE (0.001781) and the
highest composite score (0.972). Once the autoencoder hidden
layers were widened, the Medium AE reached test MSE values
of 0.006123, close to the VAE at 0.004352 and far below
the narrow Simple AE at 0.169234. PCA was at a good level, with a
value of 0.002354. This is the expected outcome given the architecture
change, and it confirms the reasoning from the design section: capacity,
not the bottleneck width, was the binding constraint.

\begin{table}[htbp]
    \caption{Test reconstruction error, Latent Score Quality, and
    the highest ranked feature for each model. Composite Selection
    Score uses Equation~(\ref{eq:composite}) with $\alpha = 0.5$.}
    \begin{center}
    \begin{tabular}{lllcl}
        \toprule
        \textbf{Model} & \textbf{MSE} & \textbf{Composite} &
        \textbf{Latent Score} & \textbf{Top 1} \\
        & & \textbf{Score} & \textbf{Quality} & \textbf{Feature} \\
        \midrule
        Simple AE        & 0.169234 & 0.407 & 0.814 & pace\_min\_km \\
        Medium AE        & 0.006123 & 0.892 & 0.810 & pace\_min\_km \\
        \textbf{Deep AE} & \textbf{0.001781} & \textbf{0.972} & \textbf{0.943} & \textbf{average\_hr} \\
        PCA              & 0.002354 & 0.927 & 0.858 & average\_hr \\
        VAE              & 0.004352 & 0.915 & 0.845 & final\_cadence \\
        \bottomrule
    \end{tabular}
    \label{tab:model_comparison}
    \end{center}
\end{table}

\begin{figure}[htbp]
    \centerline{\includegraphics[width=0.9\columnwidth]{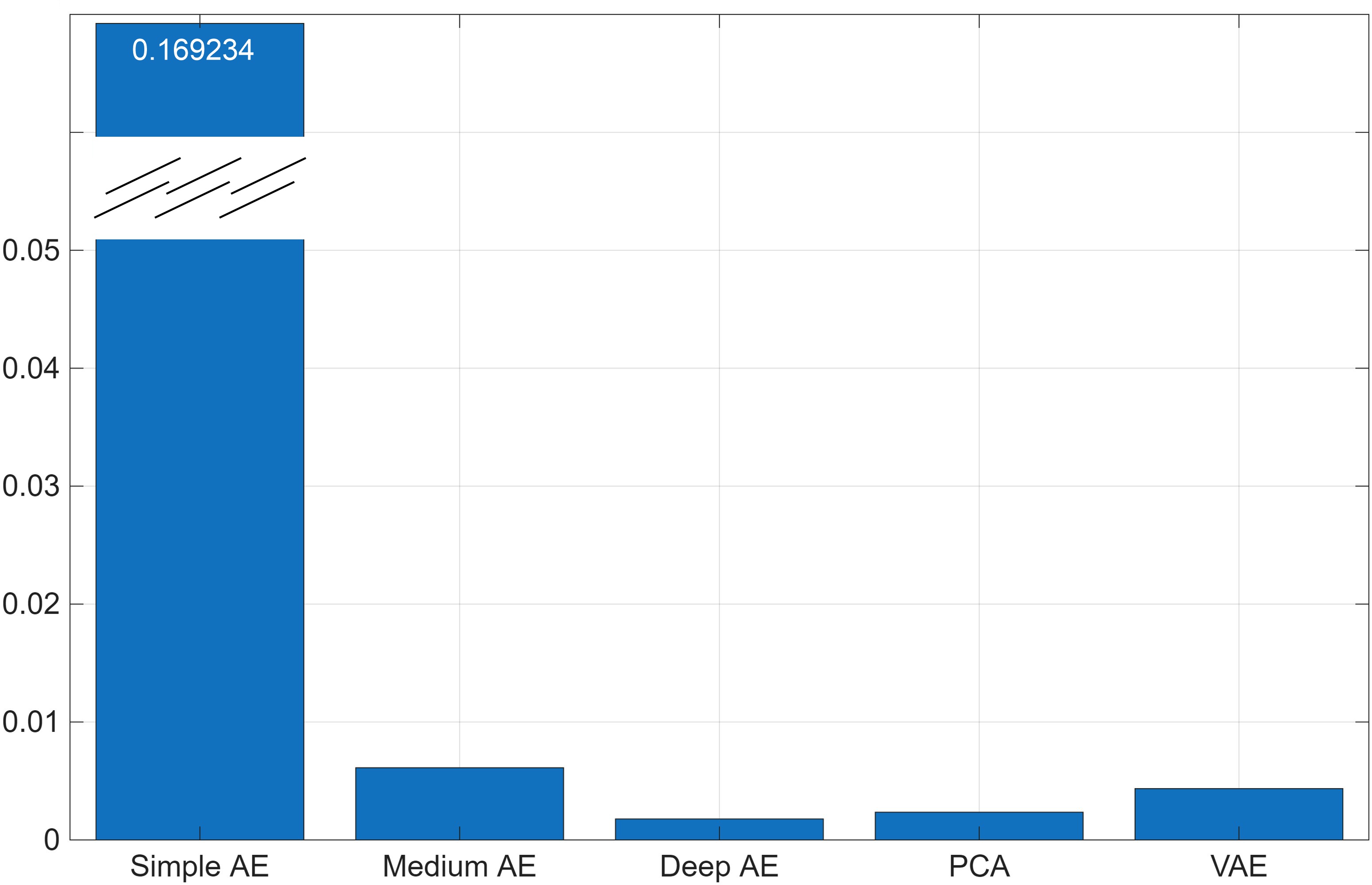}}
    \caption{Reconstruction MSE across the five models on the held out
    test set.}
    \label{fig:mse_comparison}
\end{figure}

The composite ranking is: Deep AE (0.972), PCA (0.927), VAE (0.915), Medium AE (0.892), and Simple AE (0.407). These results show
that MSE alone is an insufficient criterion for unsupervised
physiological modeling. PCA preserves linear variance and reconstructs
inputs accurately, while the nonlinear autoencoders expose latent
structure that may differ in physiological interpretation even when
their reconstruction error is comparable \cite{b3}. To check whether the small composite score gap between PCA and the best autoencoder is robust...

\begin{figure}[htbp]
\centerline{\includegraphics[width=\columnwidth]{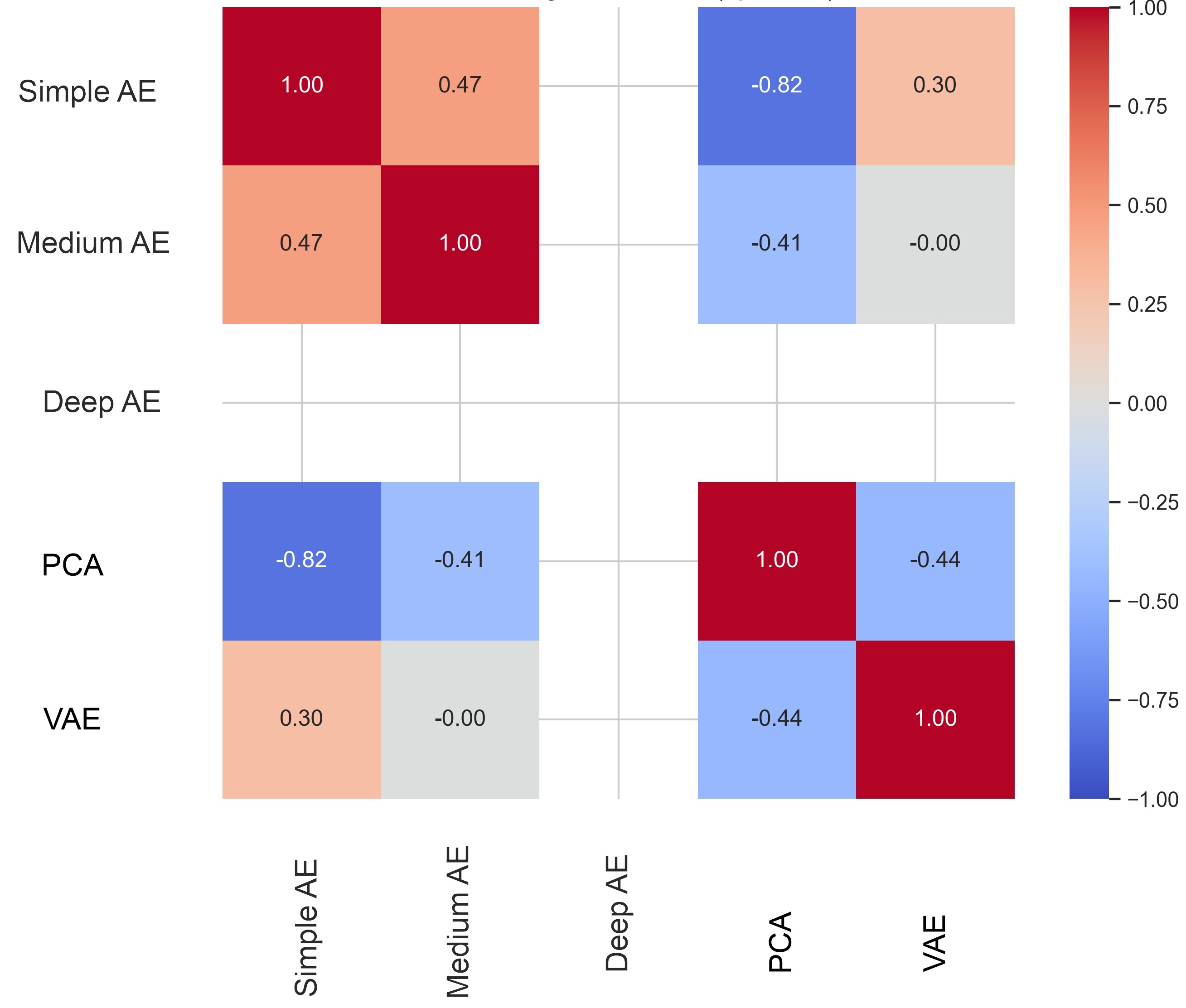}}
\caption{Spearman correlation matrix of latent scores produced by the five evaluated models.}
\label{fig:model_agreement}
\end{figure}
Consensus analysis and bootstrap validation consistently identified the same three dominant features. Running pace, aerobic decoupling, and average heart rate each reached a top 3 inclusion probability of 1.000 in both correlation based and perturbation based rankings. This makes it unlikely to be a split specific artifact, and the same trio appeared under both PCA and the autoencoders, suggesting these variables define the dominant variance structure of the dataset regardless of the compression mechanism.

\subsection{Comparison with a Linear Baseline}
To quantify the contribution of nonlinear compression, the Deep AE latent score was compared against a simple linear baseline,
\begin{equation}
    \text{Score}_{\text{base}} = w_1 \cdot \overline{\text{pace}} + w_2 \cdot \overline{\text{HR}}
\end{equation}
where $\overline{\text{pace}}$ and $\overline{\text{HR}}$ are standardized pace and average heart rate, and $w_1 = w_2 = 0.5$. This represents the simplest plausible manual combination a coach might apply. The Deep AE latent score was strongly and inversely related to this baseline (Spearman $\rho = -0.850$, Kendall $\tau = -0.681$), meaning the two scores are not interchangeable. The baseline prioritizes average heart rate, pace, and cadence, while the Deep AE prioritizes average heart rate, pace, and aerobic decoupling, indicating that the nonlinear model preserves decoupling information that the linear combination misses.

\subsection{Physiological Validation}
Representative model athlete profiles show a clear physiological separation between high ranked and low ranked groups. Leading athletes typically exhibit a pace in the range of 2.2 to 2.6 min/km, near zero aerobic decoupling, and average heart rates around 125 to 134 bpm. Lower ranked athletes frequently show substantially slower paces (12 to 15 min/km) and, in some cases, pronounced aerobic decoupling (one session reached approximately 46.9\%), consistent with marked cardiovascular drift. These contrasts support the latent score's ability to rank athlete profiles along meaningful physiological axes.

\subsection{Practical Implementation}
Both the Deep AE and PCA support deployment in a backend that processes raw FIT files and assigns athletes to contextual performance bands. Scores below 20 indicate early stage profiles; scores above 80 indicate strong profiles, giving coaches a concise summary without requiring manual inspection of telemetry tables. The latent score framework can help separate biomechanical output from temporary cardiovascular strain during heavy training blocks \cite{b11}, supporting tapering decisions before competition. The same workflow could extend to mobile tools for amateur runners and may facilitate early identification of overreaching before performance declines become clinically significant \cite{b15}.

\section{Discussion}
Deep AE achieves the highest composite score. The original Medium AE, with a single hidden layer of four units, reconstructed poorly. Widening it to sixteen units and giving the encoder brings both models close to PCA and the VAE. The VAE and PCA compresses to a single latent dimension, yet it reconstructs well, shows that the one dimensional bottleneck is not the binding constraint. The results match domain physiology findings \cite{b13}, obtained without any labeled performance outcomes or assigned by expert weights. This is meaningful for practical use, since the framework requires no domain knowledge at initialization yet recovers results that match domain knowledge when evaluated. The strong negative latent score correlation between Simple AE and PCA ($\rho = -0.65$) is a practical warning. Two models that both optimize reconstruction on the same dataset can still rank athletes in almost opposite orders. Without the agreement matrix, a practitioner selecting between them on MSE alone would not detect this problem.

\section{Limitations}
Several limitations bound the interpretation of these results. First, the dataset originates from a single training platform, which constrains external validity with respect to populations using different devices, training cultures, or geographic conditions. Establishing external validity would require rerunning the full pipeline on exports from a second device ecosystem, such as Garmin or Polar files, and confirming that pace, aerobic decoupling, and average heart rate remain the dominant features. Second, the absence of ground truth labels such as $\text{VO}_2\text{max}\;$ measurements or verified race results means the latent score can be validated only against physiological face validity, not against a criterion standard. Third, the equal composite weight ($\alpha = 0.5$) was chosen by convention; the sensitivity analysis in Table~\ref{tab:alpha_sensitivity} shows that Deep AE ranks first across $\alpha \in \{0.3, 0.5, 0.7\}$, so the conclusion is not strongly sensitive to this choice.

\section{Conclusions}
This paper described an unsupervised framework for compressing nine dimensional runner telemetry into a single interpretable performance score. Five dimensionality reduction models were evaluated on a composite criterion that combines reconstruction accuracy and latent score interpretability. Deep AE achieved the highest composite score and the lowest reconstruction error. This framework can be generalized to any unsupervised setting that requires a compact representation to be both accurate and interpretable. Bootstrap validation confirmed that pace, aerobic decoupling, and average heart rate are the dominant latent score drivers, and their stability is very high. Each reached a top 3 inclusion probability of 1.000 across all resamples. This result is consistent with established sports physiology \cite{b13} and was obtained without any domain knowledge at initialization. Future work will include supervised validation against race results or laboratory physiological parameters, cross platform generalization analysis, and temporal modeling with sequential architectures such as LSTM networks \cite{b16} or Transformers to capture within session fatigue dynamics that session aggregate features cannot represent.

\end{document}